\documentclass[final,5p,times,twocolumn,authoryear]{elsarticle}

\usepackage{amssymb}
\usepackage{stfloats}
\usepackage{xcolor,pifont}
\usepackage{tikz}
\usetikzlibrary{shapes.geometric, arrows}
\usepackage[english]{babel}
\usepackage{csquotes}

\usepackage{amsmath}
\usepackage{soul}

\newcommand{\datasetB}{\textit{Sparse-LeukemiaAttri}}
\newcommand{\datasetA}{\textit{LeukemiaAttri}}
\newcommand{\ackname}{Acknowledgement}

\journal{Medical Image Analysis}

\begin{document}

\begin{frontmatter}



\title{
Leveraging Sparse Annotations for Leukemia Diagnosis on the Large Leukemia Dataset}



\author[label1]{Abdul Rehman} 

\affiliation[label1]{organization={Intelligent Machine Lab, Information Technology University},
            addressline={ Lahore, Pakistan}} 
\author[label1]{Talha Meraj} 

\author[label2]{ Aiman Mahmood Minhas} 

\affiliation[label2]{organization={Department of Hematology, Chughtai Lab },
           addressline={ Lahore, Pakistan}} 
\author[label2]{Ayisha Imran} 


\author[label1]{Mohsen Ali} 


\author[label1]{Waqas Sultani} 
\author[label3]{Mubarak Shah}
\affiliation[label3]{organization={Center for Research in Computer Vision, University of Central Florida},
            addressline={United States}} 


\begin{abstract}


{Leukemia is the $10^{th}$  most frequently diagnosed cancer and one of the leading
causes of cancer-related deaths worldwide.  Realistic analysis of leukemia
requires white blood cell (WBC) localization, classification, and morphological assessment.
Despite deep learning advances in medical imaging, leukemia analysis lacks a large, diverse multi-task dataset, while existing small datasets lack domain diversity, limiting real-world applicability. 
 To overcome dataset challenges, we present a large-scale WBC dataset
named ‘Large Leukemia Dataset’ (LLD) and novel methods for detecting WBC with their attributes.}
Our contribution here is threefold. 
First, we present a large-scale Leukemia dataset collected through Peripheral Blood Films (PBF) from {$48$} patients, through multiple microscopes, multi-cameras, and multi-magnification. 
To enhance diagnosis explainability and medical expert acceptance, each leukemia cell is annotated at 100x with 7 morphological attributes, ranging from Cell Size to Nuclear Shape.
Secondly, we propose a multi-task model that not only detects WBCs but also predicts their attributes, providing an interpretable and clinically meaningful solution. 
Third, we propose a method for WBC detection with attribute analysis using sparse annotations. This approach reduces the annotation burden on hematologists, requiring them to mark only a small area within the field of view. Our method enables the model to leverage the entire field of view rather than just the annotated regions, enhancing learning efficiency and diagnostic accuracy.
 From diagnosis explainability to overcoming domain-shift challenges, the presented datasets can be used for many challenging aspects of microscopic image analysis. The datasets, code, and demo are available at: {https://im.itu.edu.pk/sparse-leukemiaattri/}.
\end{abstract}




\end{frontmatter}





\section{Introduction}
\label{sec:introduction}

 {Leukemia is among the leading causes of cancer-related deaths worldwide, with an estimated 487,000 new cases and 305,000 deaths reported in 2022 \cite{bray2024global}. 
This disease disproportionately affects people under the age of 40, with children being particularly impacted \cite{creutzig2018acute}. The diagnosis of leukemia requires an expert hematologist with extensive experience, capable of analyzing peripheral blood slides employing costly medical equipment. {Furthermore, morphological examination remains a cornerstone of leukemia diagnosis, providing accessible, low-cost, and widely available insights into hematologic malignancies} \cite{wintrobe2009wintrobe}.  Given the prerequisites for the diagnosis and prognosis of leukemia, many remote regions in underdeveloped countries experienced higher mortality rates \cite{rugwizangoga2022experience}. However, early detection of leukemia enables timely and accurate treatment, which can significantly improve survival rates.
For leukemia diagnosis, hematologists examine the peripheral blood film (PBF) of patients, a process that is often tedious, error-prone, and requires significant expertise. The limited availability of experts and costly diagnostic resources hinders early, precise, and timely detection of leukemia, particularly in resource-constrained areas. Furthermore, leukemia treatment requires repetitive blood sample preparation and detailed analysis, requiring frequent access to experts. 
AI-driven solutions are vital to overcome the low accessibility of experts. Such a system should be able to work across multiple microscopes and digitization scenarios, and to augment hematologists, it should provide explainable information to hematologists. Unfortunately, existing leukemia datasets ~\cite{scotti2005automatic,labati2011all,rezatofighi2011automatic,matek2019single,kouzehkanan2022large,bodzas2023high}  suffer from many limitations, including the absence of localization annotations of white blood cells, the lack of associated morphological attributes, and a limited number of images \cite{gao2023childhood}. {Furthermore, there is no established multi-task learning (MTL) \cite{he2024multi,xu2023regional,koyuncu2020deepdistance} approach that simultaneously detects white blood cells and predicts their morphological attributes.}

In addition to these limitations, these supervised algorithms (\cite{sun2021sparse, tian2019fcos, zhang2022dino, ultralytics2021yolov5, wollmann2021deep}) require every cell in each image to be annotated, adding a significant manual burden.
In the practical scenario, a hematologist uses only a small portion of the acquired image to perform the required analysis and annotation. 
Thus majority of the cells remain unannotated. On the other hand, requiring annotation of complete blood films not only increases time and cost but also limits the number of blood films that could be annotated \footnote{Note that multiple image patches are extracted at the specified resolution}.
Employing existing object detection strategies over sparsely annotated datasets results in subpar performance since these methods are unable to use the unlabeled parts of slides. Complete annotation of a blood film in the monolayer takes around 200 minutes—eight times longer than the 25 minutes needed for sparse annotation.

To overcome dataset challenges, we present a large-scale WBC dataset named `large leukemia dataset' (LLD), that consists of two subsets `LeukemiaAttri'\footnote{This subset was published in \textit{`A Large-scale Multi Domain Leukemia Dataset for the WBC Detection with Morphological Attributes for Explainability'} at MICCAI 2024 \cite{rehman2024large}, and this journal paper serves as its extension.}, capturing cell localization, classification, and morphological attribute prediction tasks; and `Sparse-LeukemiaAttri', for the object-detection and attribute prediction based on sparsely annotated dataset task. An important highlight of the \datasetA~ is that it has been collected through two microscopes from different price ranges, one high-cost (HCM) and one low-cost (LCM), at three magnifications (100x, 40x, 10x) using various sensors: a high-end camera for HCM, a mid-range camera for LCM, and a mobile phone camera for both microscopes. 
In this collection, using high-resolution HCM (100x), experienced hematologists annotated 10.3k WBC of 14 types, having 55k morphological labels from 2.4k images of several PBS leukemia patients.
The data set collection procedure is illustrated in Fig \ref{fig:Leukemiaattri_data} and further details are in section \ref{sec:dataset1}. 
\begin{figure}[!h]
 \centering
    {\includegraphics[width=1\linewidth]{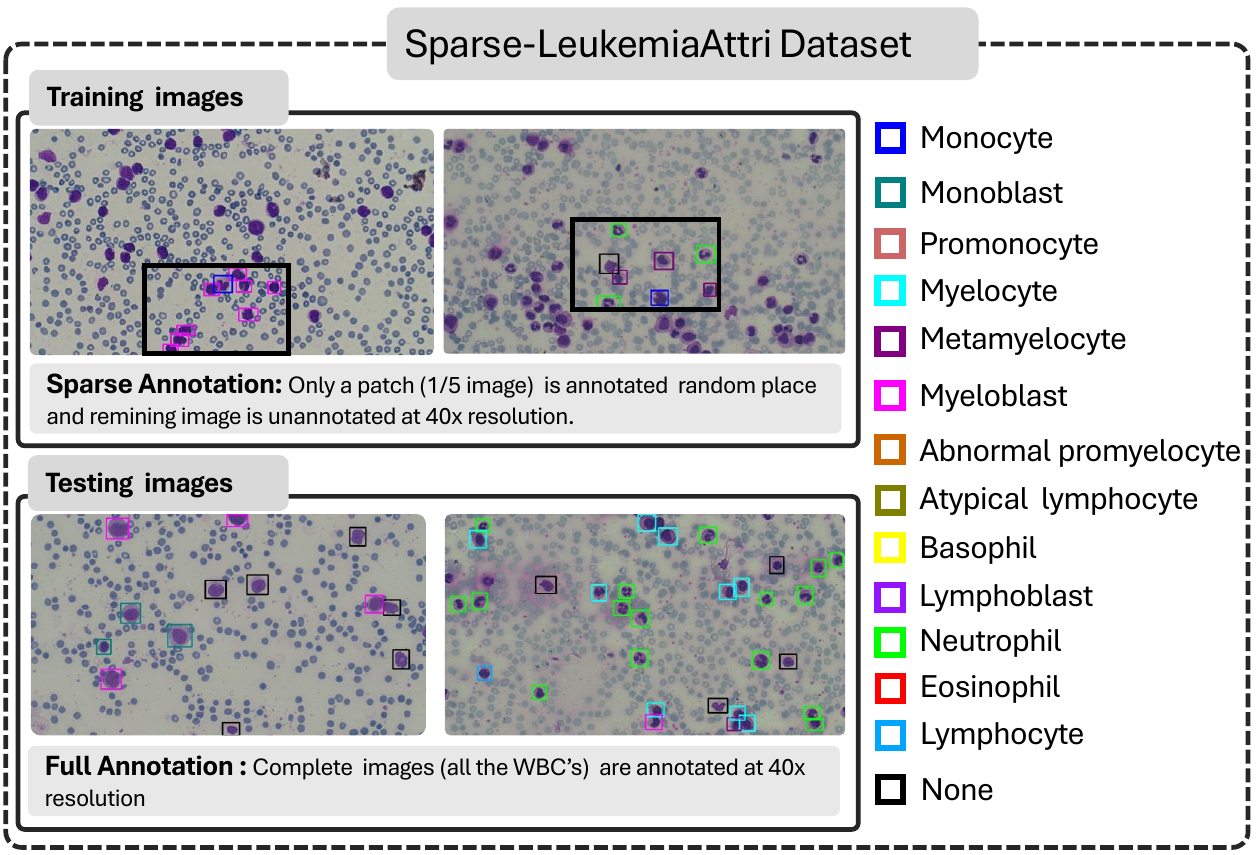}}
     \caption{Sparse-LeukemiaAttri Dataset: In the training set, images are sparsely annotated, with only a random $(1/5)^{th}$ of each image labeled. For the testing set, the entire image is fully annotated to allow for comprehensive evaluation.}

    \label{fig:Sparse-LeukemiaAttri}
\end{figure}
In conjunction with these datasets, we propose two models named leukemia attribute detector (AttriDet) and sparse Leukemia attribute detector (SLA-Det). 
For the task of detection of white blood cells (WBCs) along with the prediction of morphological attributes, we integrated a lightweight head, named AttriHead, into the existing object detector. AttriHead employs multiscale features from the feature backbone used in object detection. This not only enhances computational efficiency but also enables the backbone to become more informative through MTL. 
For the sparsely-supervised object detection, we propose a novel algorithm called SLA-Det. Note that in our case, on average, only 20\% of the image is annotated shown in Figure \ref{fig:Sparse-LeukemiaAttri}. 
To effectively use the unannotated region information, SLA-Det incorporates region-guided masking and pseudo-label selection based on multiple criteria, ensuring robust model performance. We believe that these extensions align our dataset and framework more with clinical practices and enhance the reliability of automated leukemia detection.
Our extensive experimental results and discussion validate the proposed ideas and framework. Our contributions are given below.

\begin{itemize}
    \item \textbf{Large leukemia dataset (LLD)} consists of LeukemiaAttri and Sparse-LeukemiaAttri. 
    \textbf{LeukemiaAttri:} A large-scale, multi-domain dataset of 2.4K microscopic images captured using microscopes of varying costs, two cameras, and three resolutions (10x, 40x, 100x). Each image includes WBC localization and morphological attribute annotations for an explainable diagnosis.  \textbf{Sparse-LeukemiaAttri:} A 40x FoV dataset with sparsely annotated training data (1,546 images) and fully annotated test data (185 images).
    \item  Two detection models are proposed:
    \begin{itemize}
        \item \textbf{AttriDet} is a fully supervised detector designed to identify WBCs and their morphological attributes.
        \item \textbf{SLA-Det} is a sparse WBC detector that enhances pseudo-label selection by incorporating problem-specific constraints, including size, objectness score, and class label entropy.
    \end{itemize}
    \item Key results: Using sparse annotations, SLA-Det can process slides at 40x resolutions while maintaining accuracy, unlike AttriDet, which only worked on 100x resolution; thus, it can process slides 4x faster than \textbf{AttriDet}. SLA-Det outperforms state-of-the-art sparse detection methods by 34.7 $mAP@_{50}$, demonstrating superior efficiency and performance.


    
\end{itemize}

The work emphasizes domain diversity, diagnostic interpretability, and efficient sparse annotation strategies for real-world leukemia analysis.

\section{Related Work}
Our dataset includes densely labeled patches and sparsely labeled large images, allowing for both fully supervised Object Detection (OD) and Sparsely Annotated Object Detection (SAOD). Below is a brief review of these methods that highlights their relevance to WBC detection.\\
\noindent\textbf{Supervised WBC detection: }
Deep learning (DL) techniques have been extensively applied to WBC classification \cite{jung2019w,wu2023feature,wang2022deep,gehlot2020sdct,gehlot2021cnn}. Huang \textit{et al.} \cite{huang2020attention} implemented a two-step approach for their end-to-end solution: first, detecting the WBCs, and then performing classification after cropping the detected cells. To classify single-cell images, Saidani \textit{et al.} \cite{saidani2024white} optimized a CNN model and employed multiple pre-processing techniques, such as color conversion and image histogram equalization. 
Similarly, the authors in \cite{thanh2018leukemia} classified two types of WBCs from the ALL-IDB data set \cite{thanh2018leukemia}. G Liang et al. \cite{liang2018combining} developed a combined model of convolutional neural networks (CNN) and recurrent neural networks (RNN) to classify four types of WBCs using the BCCD dataset \cite{BCCD}.
 To detect WBCs in whole microscopy images, the authors in ~\cite{han2023one} gave a lighter deep-learning CNN model and adapted the YOLOX-nano \cite{ge2021yolox} baseline to identify the types of WBC. Their method enhances the overall model architecture by improving Intersection over Union (IoU) loss.
{A blood film and bone marrow aspirates-based multiple instance learning for leukocyte identification approaches have been proposed to classify different types of Leukemia ~\cite{manescu2023detection}. For image preprocessing, they mainly applied threshold methods.}\\ 
\noindent\textbf{Sparsely annotated object detection: }
{To tackle the challenge of object detection in sparsely annotated datasets, authors in \cite{suri2023sparsedet} proposed a method that integrates supervised and self-supervised learning into their framework. The approach involves extracting features from an image and its multiple augmented versions, which are then processed through a shared region proposal network. A positive pseudo-mining technique is introduced to effectively filter out labeled, unlabeled, and foreground regions. This results in separate losses for labeled regions, supervised loss and self-supervised loss, ensuring consistent feature extraction across both original and augmented views. This combined method enhances the model's generalization ability and improves object detection performance. Similarly, Z. Wu \textit{et al.} \cite{wu2018soft} used a soft sampling technique to mitigate the impact of missing annotations. Their approach applies the Gompertz function to adjust gradient weights based on the overlap between positive instances and regions of interest. Niitani et al. \cite{niitani2019sampling} propose a part-aware learning approach that trains a model twice, using pseudo labels from the first model to guide the second. By leveraging spatial relationships, it selectively ignores classification loss for part categories within subject categories, improving training and performance with sparse annotations. To generate labels for sparsely annotated datasets, the co-mining approach is applied by the authors of \cite{wang2021co}. It involves constructing a Siamese detection network with two branches, each generating pseudo-label sets using a co-generation module.  
It is subsequently used as guided signals to the model, leveraging augmentation techniques to generate diverse pseudo-label sets. 
Finally, a co-generation module converts predictions from one branch into pseudo-labels for the other, facilitating complementary supervision between the network branches. Note that previous benchmarks, such as \cite{suri2023sparsedet, wang2021co}, implemented sparse object detection algorithms on self-created datasets. In contrast, our dataset consists of small patches annotated with detailed morphological features and patch localization. In addition, we introduce a multi-headed WBC detection method tailored for such data. 
\section{Dataset}
\label{sec:dataset1}
Leukemia comprises four main types: acute lymphocytic leukemia (ALL), chronic lymphocytic leukemia (CLL), acute myeloid leukemia (AML), and chronic myeloid leukemia (CML) \cite{bray2024global}. These types are categorized into two major lineages: myeloid (AML and CML) and lymphoid (ALL and CLL). Each leukemia type exhibits distinct abnormalities, including increased myeloblasts in CML,
atypical lymphocytes in CLL, elevated myeloblasts in AML, and excessive lymphoblasts in ALL. However, as shown in Table \ref{tab:EXISTING LEUKEMIA DATASETS}, most existing white blood cell (WBC) datasets were collected using a single microscope, lacked bounding boxes and morphological details, and often included either only healthy individuals or patients with a single leukemia type.}

\begin{table*}[h]
  \centering
    \caption{Comparison of the proposed dataset with existing leukemia datasets.  }  
    \label{tab:EXISTING LEUKEMIA DATASETS}

\resizebox{\linewidth}{!}{
\begin{tabular}{|l|l|c|c|c|c|l|c|c|}
    \hline
    &&Multi.&Multi. Cells & Bounding &Multi.&No. of & WBC& Morphology \\
    Dataset&Type&Micro.&in image&Box&Res.&WBC's&Classes&\\
    \hline
     IDB \cite{scotti2005automatic} & ALL&\ding{56}
     & \textcolor{orange}{\ding{52}} 
&\textcolor{orange}{\ding{52}}&\ding{56}&510 (LB)&2& 
        \ding{56}\\
        IDB2 \cite{labati2011all} & ALL & \ding{56}& 
        \ding{56}
         & \ding{56} &\ding{56}&260&2& 
        \ding{56}\\
        LISC \cite{rezatofighi2011automatic}& Normal & \ding{56}
        &\ding{56} 
        & \ding{56}&\ding{56}&  250&5& 
        \ding{56}\\
        AML Matek\cite{matek2019single} & AML & \ding{56}
        & \ding{56}
        & \ding{56}&\ding{56}&18,365&15& 
        \ding{56} \\
       Raabin \cite{kouzehkanan2022large}& Normal & \ding{56} & \textcolor{orange}{\ding{52}}& \textcolor{orange}{\ding{52}}&\ding{56}& 
     17,965& 5& 
        \ding{56}\\
        HRLS \cite{bodzas2023high}   & Multi. & \ding{56} &\textcolor{orange}{\ding{52}}
       & \ding{56}&\ding{56} &16,027&9& 
        \ding{56}\\
 {Acevedo}~\cite{acevedo2020dataset} & {Normal} & {\ding{56}} & {\ding{56}}
       &{ \ding{56}}&{\ding{56}} &{17,092}&{10
       }& 
       {\ding{56}}\\

       AML Hehr \cite{hehr2023morphological} & AML & \ding{56} &\ding{56}
       & \ding{56}&\ding{56} &81,214&4& 
       \ding{56}\\
         WBCAtt \cite{tsutsui2023wbcatt}   & Normal & \ding{56} &\ding{56}
       & \ding{56}&\ding{56} &10,298&5& 
        \textcolor{orange}{\ding{52}}\\
       
     \textbf{\datasetA} \footnotemark &Multi.&\textcolor{orange}{\ding{52}}
     &\textcolor{orange}{\ding{52}}&\textcolor{orange}{\ding{52}}&\textcolor{orange}{\ding{52}}
  &\textbf{88,294}&14& 
        \textcolor{orange}{\ding{52}}\\
    \hline
  \end{tabular}
  }
\end{table*}
 \footnotetext{The details LeukemiaAttri is explained in Table \ref{tab:dataset_detail}}

\subsection{Limitations of existing WBC datasets}

The details of publicly available datasets on WBC and leukemia are shown in Table \ref{tab:EXISTING LEUKEMIA DATASETS}, which highlights various limitations such as:
1)  The ALL-IDB dataset \cite{scotti2005automatic} contains only 510 WBCs from patients with Acute Lymphocytic Leukemia (ALL) and includes only cell centroid information.
2) In version 2 of the ALL IDB \cite{labati2011all}, each image contains only a single WBC, with the cell's status labeled accordingly. 
3) The LISC \cite{rezatofighi2011automatic}  dataset contains a smaller number of WBCs with five classes, such as basophil, eosinophil, lymphocyte, monocyte, and neutrophil.
{4) The entire dataset of AML Matek \cite{matek2019single} and Acevedo \cite{acevedo2020dataset} contains single-cell images that come under the umbrella of AML and normal individuals.} 
5) Only Raabin \cite{kouzehkanan2022large}, out of all the aforementioned existing datasets, has complete bounding box-level annotation with single-cell images, but the dataset consists entirely of normal individuals, with over 60\% of the images containing only a single cell in the FoV.
Furthermore, the Raabin-WBC collection contains images from two separate cameras and two different microscopes, without maintaining any correspondence across microscopes. 
6) The HRLS \cite{bodzas2023high} dataset is captured using a single microscope and covers only two types of Leukemia and nine types of WBCs.
7) WBCAtt dataset  \cite{tsutsui2023wbcatt} is the improved version of \cite{acevedo2020dataset}, where annotations of attributes are added for each WBC.  The dataset \cite{acevedo2020dataset} lacks the abnormal WBCs and is captured using a single microscope only.


\begin{figure*}[h]
 \centering
   {\includegraphics[width=1\linewidth]{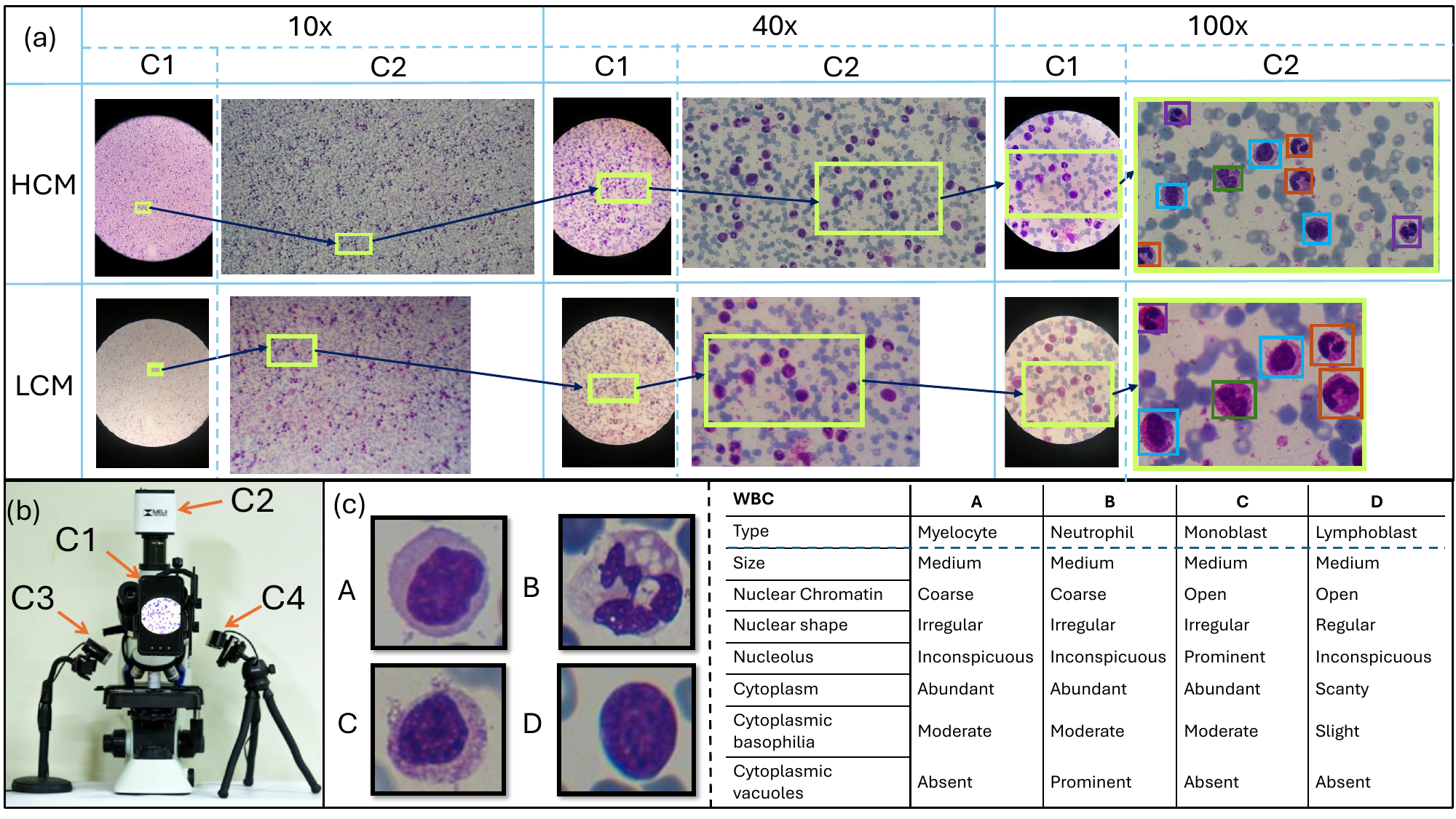}}
   \caption{a) Illustrates the image acquisition procedure utilizing a standard mobile camera (C1), a premium camera for high-cost microscopy (HCM), and a mid-range camera (C2) for low-cost microscopy (LCM) at various resolutions (100x, 40x, and 10x). The images are obtained with both high and low-cost microscopes. b) Displays our microscope set up with four cameras, C1 to C4. C1 and C2 capture the field of view of the microscope lens, and C3 and C4 capture the stage scale. c) This table presents various types of white blood cells (WBC) and their distinctive morphological features.}
    \label{fig:Leukemiaattri_data}
\end{figure*}
\subsection{Our dataset collection procedure}

To understand leukemia prognosis, we consulted healthcare professionals to finalize key WBC types and their morphology \cite{wintrobe2009wintrobe}. To improve deep learning across modalities, we propose capturing images of four leukemia PBF types from multiple patients at various magnifications using available resources. Capturing images at varying resolutions enhances deep learning generalization, enabling low-cost, automated leukemia diagnostics. 
{The dataset comprises 48 Peripheral Blood Films (PBFs) collected from distinct leukemia patients (18 ALL, 22 AML, 2 CLL, 4 CML, and 2 APML), with patient-level splits applied to prevent data leakage and ensure robust evaluation.} 
Images are captured using two microscopes: the high-cost Olympus CX23 and the low-cost XSZ-107BN, along with three cameras: the HD1500T (HCM), ZZCAT 5MP (LCM), and the Honor 9x Lite mobile camera. The high-end camera (ZZCAT 5MP) is 47 times more expensive than the low-end camera (XSZ-107BN), while the HCM is 17 times more costly than the LCM. Although the LCM is more affordable, it sacrifices visual detail and has a smaller field of view. Note that capturing the same WBC across multiple microscopes and resolutions is highly challenging due to several factors: (a) Locating the same WBC patch on the peripheral blood film (PBF) across different microscopes and resolutions is inherently difficult \cite{sultani2022towards},
(b) Microscope stage scales are not calibrated uniformly across different microscopes. 
(c) Even within the same microscope, poor alignment leads to resolution inconsistencies, and
(d) Viewing a patch at 100x requires immersion oil, whereas 10x and 40x provide better visibility without it. This necessitates repeatedly adding and removing oil when capturing the same patch at multiple resolutions. 

We began the capturing process by setting the field of view (FoV) at 10x with approximately 20\% overlap while keeping a fixed x-axis microscope stage scale. We then replicated the same procedure for the 40x magnification at an identical scale. At 100x magnification, we captured the corresponding FoV without overlap, ensuring each WBC was distinctly represented. 
To achieve this, we used an in-house developed software solution that enables simultaneous FoV image capture from mobile (C1), microscope-specific cameras (C2), and stage-capturing cameras (C3 and C4). For cross-microscope comparison, we first calibrated the scale stages of both microscopes and then repeated the process for the LCM.

\subsection{LeukemiaAttri dataset} 
\label{Sec:Dataset}
To overcome the above-mentioned dataset limitations and provide a benchmark for training deep learning models, we present a large-scale dataset called the LeukemiaAttri dataset. 
The LeukemiaAttri dataset comprises 14 types of WBCs. The details of each type in a subset are presented in Table \ref{tab:classes_details}, where the \enquote{None} category includes artifacts, bare cells, and cells that were challenging for hematologists to identify.
In total, the data set contains 28.9K images (2.4K images $\times$ 2 microscopes $\times$ 3 resolutions $\times$ 2 camera types), representing 12 subsets of images captured using low- and high-cost microscopes at three magnification levels. The detail of the LeukemiaAttri dataset is explained in Table \ref{tab:dataset_detail}.

\begin{table*}[h]
\centering
\small
\renewcommand{\arraystretch}{1.1}
\caption{LeukemiaAttri dataset details with its multi-domain subsets}
\label{tab:dataset_detail}
\resizebox{\textwidth}{!}{ 
\begin{tabular}{|c|l|l|c|l|c|c|c|}
\hline
Sr. \# & Dataset Name & Microscope & Resolution & Camera & Image Count & WBC Count & Artifacts \\
\hline
1 & HCM\_100x\_C1 & OlympusCX23  & 100x & Honor 9X lite  & 2.4k & 7.8k & 2.5k \\
2 & HCM\_100x\_C2 & OlympusCX23 & 100x & HD1500T  & 2.4k & 7.8k & 2.5k \\
3 & HCM\_40x\_C1 & OlympusCX23 & 40x & Honor 9X lite & 2.4k & 7.8k & 2.5k \\
4 & HCM\_40x\_C2 & OlympusCX23 & 40x & HD1500T & 2.4k & 7.8k & 2.5k \\
5 & HCM\_10x\_C1 & OlympusCX23 & 10x & Honor 9X lite & 2.4k & 7.8k & 2.5k \\
6 & HCM\_10x\_C2 & OlympusCX23 & 10x & HD1500T & 2.4k & 7.8k & 2.5k \\
7 & LCM\_100x\_C1 & XSZ\_107BN & 100x & Honor 9X lite & 2.4k & 4.8k & 1.1k \\
8 & LCM\_100x\_C2 & XSZ\_107BN & 100x & ZZCAT-5MP & 2.4k & 4.8k & 1.1k \\
9 & LCM\_40x\_C1 & XSZ\_107BN & 40x & Honor 9X lite & 2.4k & 7.8k & 2.5k \\
10 & LCM\_40x\_C2 & XSZ\_107BN & 40x & ZZCAT-5MP & 2.4k & 7.8k & 2.5k \\
11 & LCM\_10x\_C1 & XSZ\_107BN & 10x & Honor 9X lite & 2.4k & 7.8k & 2.5k \\
12 & LCM\_10x\_C2 & XSZ\_107BN & 10x & ZZCAT-5MP & 2.4k & 7.8k & 2.5k \\ 
\hline
\multicolumn{8}{|c|}{\textbf{Sparse-LeukemiaAttri}} \\
\hline
13 & HCM\_40x\_C2\_Sparse & OlympusCX23 & 40x & HD1500T  & 1.7k & 7.6k & 2.0k \\
\hline
\end{tabular}
} 
\end{table*}

\begin{table*}[h]
\centering
\small
\renewcommand{\arraystretch}{1.1}
\caption{WBC types and their distribution in a single subset of the LeukemiaAttri (\textbf{A}) and Sparse-LeukemiaAttri (\textbf{B}) datasets.}
\label{tab:classes_details}
\resizebox{\textwidth}{!}{ 
\begin{tabular}{|c|l|c|c||c|l|c|c|}
\hline
Sr. \# & Type  of cell & Count in  \textbf{A} & Count in  \textbf{B} & Sr. \# & Type of cell & Count in \textbf{A} & Count in  \textbf{B} \\
\hline
1  & Monocyte & 205 & 189  & 8  & Monoblast & 484 & 332 \\
2  & Basophil & 19 & 18  & 9  & Lymphoblast & 1722 & 1588 \\
3  & Promonocyte & 346 & 335  & 10 & Myelocyte & 304 & 329 \\
4  & Neutrophil & 1293 & 1153  & 11 & Eosinophil & 85 & 95 \\
5  & Metamyelocyte & 101 & 133  & 12 & Myeloblast & 2157 & 2029 \\
6  & Lymphocyte & 322 & 282  & 13 & Atypical lymphocyte & 419 & 420 \\
7  & Abnormal promyelocyte & 276 & 298  & 14 & None & 2560 & 2375 \\
\hline
\end{tabular}}
\end{table*}

\subsubsection{Morphological attributes:}
\label{sub:morphology_detail}
To improve prognostic analysis, hematologists identified 14 distinct types of WBCs. The morphological characteristics of several types of WBC are illustrated in Figure \ref{fig:Leukemiaattri_data}
(c). For annotating the WBC types and their morphological features, hematologists meticulously reviewed all subsets of the captured images and selected the subset with the best-quality images that offered clear structural details.  This selected subset (HCM\_100x\_C2), which was captured using the high-cost microscope (\textbf{HCM}) at \textbf{100x} magnification with the HD1500T camera (\textbf{C2}), was used for the annotation process. \\
To ensure accuracy, two hematologists collaborated in annotating the cells. Figure \ref{fig:Leukemiaattri_data}
(c) shows examples of certain WBC types and their morphological attributes, such as monoblasts, monocytes, and myelocytes (A, B, C in the Figure~\ref{fig:Leukemiaattri_data}(c)) which exhibit similar morphological features due to their shared myeloid lineage. 
However,  variations are observed, especially in the presence of cytoplasmic vacuoles. 
On the other hand, lymphoblasts (Figure~\ref{fig:Leukemiaattri_data}(c): D), which belong to the lymphoid lineage, exhibit distinct morphological characteristics compared to the myeloid lineage. 
Once detailed annotations of WBC types and their attributes were obtained at 100x magnification using the HCM, we transferred these annotations to images captured at different resolutions and with the LCM using homography techniques \cite{lowe2004distinctive,fischler1981random}. The transferred annotations were manually reviewed, and any missing or incorrectly localized annotations were re-annotated. The specific counts of WBC types and their corresponding attributes for the source subset are provided in Table \ref{tab:classes_details} 
\subsection{Sparse LeukemiaAttri dataset} 
\label{Sec:Dataset}

To ensure the accuracy of morphological attributes, the LeukemiaAttri dataset was collected at 100x magnification, and then the corresponding patches from 40x and 10x were extracted while ignoring the complete FOV of 40x and 10x. 
To make efficient use of available data and to ensure practical efficacy (since 40x is the gold standard for Leukemia), in the Sparse LeukemiaAttri dataset, the complete FOV of 40x should be considered. 
\begin{figure}[!htbt]
 \centering
    \includegraphics[width=.9\linewidth]{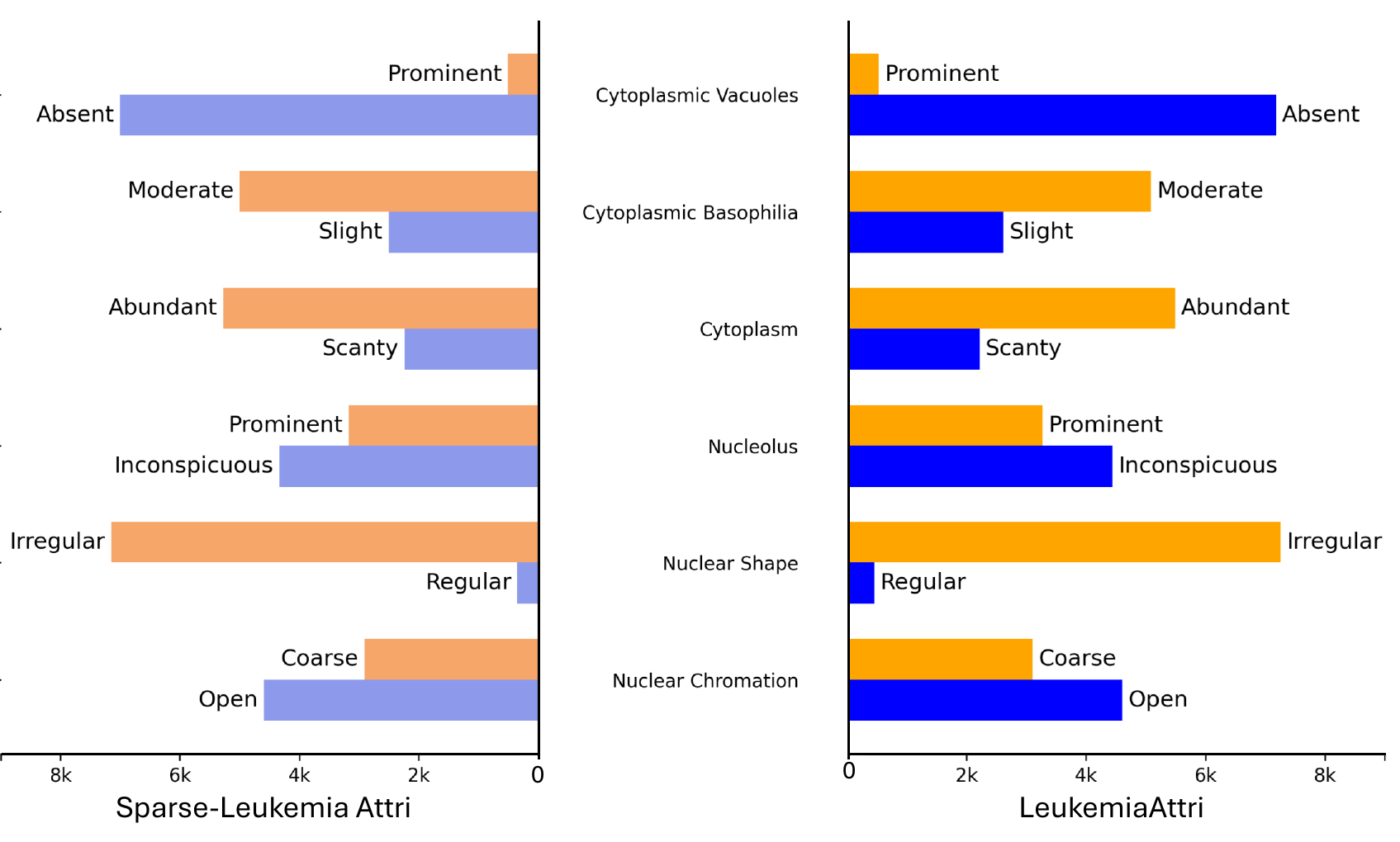}
    \caption{Attribute classes distribution of a subset from LeukemiaAttri and Sparse leukemiaAttri datasets. The distribution shows that the class imbalance in the Nuclear shape and Cytoplasmic vacuoles is much higher than the others.}
    \label{fig:dataset_detail}
\end{figure}
To address the above limitation of the LeukemiaAttri dataset,  we extend the $HCM\_40X\_C2$ (see Table 2) subset to include a complete FoV.
This new dataset is split up into a train set (1546 images) and a test-set (185 images). 
For the training set, the annotations are borrowed from the LeukemiaAttri dataset and mapped to 40x images, thus resulting in the sparse annotations and named as \datasetB. 
 {The ratio of the labeled patch to the entire field of view (FoV) at 40x, in terms of size, is 1:5.}
However, for the test set (curated at 40x resolution), each patch of the image is annotated at 100x (for better annotation quality) resolution, and these annotations are transferred to a complete FOV of 40x test set. Examples of train and test subsets are shown in Figure~\ref{fig:Sparse-LeukemiaAttri}.
Collectively, with a new extension, we have 13 subsets now as shown in  Table \ref{tab:dataset_detail}. \\
Our \datasetB~dataset is a specialized case of sparsely supervised and semi-supervised object detection datasets. Unlike traditional sparsely supervised datasets, where annotations are randomly missed, each of our images contains a fully labeled patch. Similarly, while semi-supervised approaches typically involve fully labeled and fully unlabeled sets, our dataset consists of partially annotated patches within each image, rather than entirely labeled or unlabeled images.

\section {Proposed Multi-Task Detectors}

Our goal is to detect WBC, along with their morphological attributes, from PBF images to enable a real-time, explainable diagnosis of leukemia.
To avoid a large computational cost, instead of a separate network, we extract features from the backbone of the predicted WBC and estimate the attributes of the WBC through multi-task learning. 
Furthermore, to adhere with the real-world scenario, where hematologists analyze the PBF on 40x for efficiency, the proposed WBC detector, including the attribute predictor, is trained on the complete FOV of 40x  resolution.
Since data was annotated on 100x, it maps to only a small portion in 40x, resulting in sparse annotation.
Therefore, we design a self-learning mechanism to train a multi-task object detector in a sparse setting.


\subsection{Preliminaries}
Let $\mathcal{F}(I_i)=\{\hat{\mathbf{z}}_{i,k}\}_{k=1}^{n_i}$ be the object detector that outputs $n_i$ predictions. 
Each prediction \(\hat{\mathbf{z}}_i^k =( \hat{{c}}_{i,k}, \hat{{o}}_{i,k}, \hat{\mathbf{p}}_{i,k}, \hat{\mathbf{b}}_{i,k})\) consists of objectness score $\hat{{o}}_{i,k}$, a class with maximum probability $\hat{c}_{i,k}$, predicted probability vector $\hat{\mathbf{p}}_{i,k}$, and bounding-box $\hat{\mathbf{b}}_{i,k} \in \mathbb{R}^{4}$. 
The confidence score, $\hat{p}_{i,k}$, for $k^{th}$ prediction is computed as: $max(\hat{\mathbf{p}}_{i,k}) \times \hat{o_{i,k}}$,
where $max$ returns the maximum value from the input vector.
For readability, we assume that $\hat{\mathbf{b}}_{i,k} = \{x_k, y_k, w_k, h_k\}$ and
$I_i\in \mathbb{R}^{H \times W}$ represents the input image and $\{ \mathbf{z}_{i,k} \}_{k=1}^{n_i} $ is the annotation of the input image. Let $\mathcal{D_A} = \{(I_{i},  \{\mathbf{z}_{i,k}\}_{k=1}^{n_i})\}_{i=1 }^{N_D}$ be the fully annotated dataset.
Each annotation $\mathbf{z}_i^k = (\mathbf{b}_{i,k}, \mathbf{c}_{i,k}, \mathbf{m}_{i,k})$ consists of a bounding box $\mathbf{b}_{i,k}$, a corresponding class label $\mathbf{c}_{i,k} \in \{1, \dots, C\}$, and $\mathbf{m}_{i,k}$, which is the set of morphological attributes associated with the $k^{\text{th}}$ object in the $i^{\text{th}}$ image.

Our methodology is compatible with any object detector, however, we chose YOLOv5 \cite{ultralytics2021yolov5} for our approach, since our experiments indicated (Table \ref{tab:Baseline_results}) it to have higher mAP in comparison with other recent SOTA object detectors including FCOS \cite{tian2019fcos}, Spare-Faster-RCNN \cite{sun2021sparse} and transformer-based DINO\cite{zhang2022dino}.
 
\begin{figure*}[h!]
 \centering
\includegraphics[width=1\linewidth]{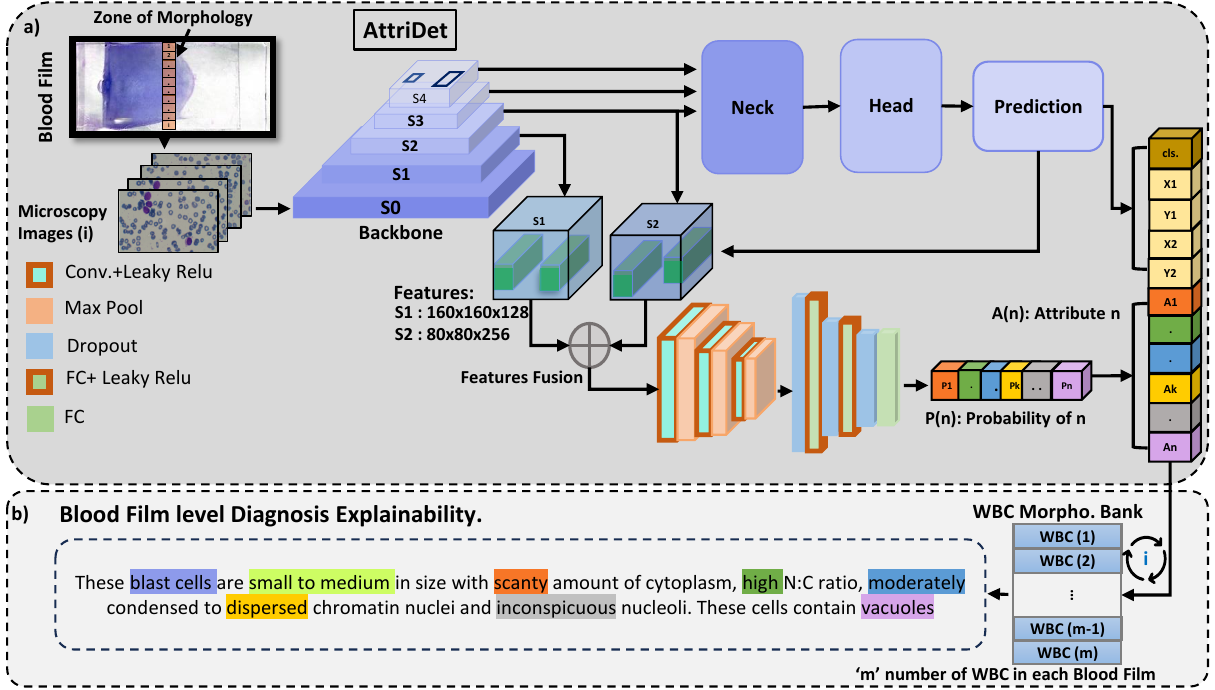}
    \caption{
     {Architecture of Attri-Det: The proposed framework for white blood cell (WBC) and morphological attribute detection processes input images using a backbone network to extract features. 
    The WBC detector is trained with standard detection losses, while the attribute head uses asymmetric multilabel classification loss.
    }
    b) Predictions from each image are stored in a morphology bank, updating the ‘Masked Text’ (shown in the different colors in the figure) with the predominant WBC type and morphology for a comprehensive blood film-level description. }
    \label{fig:AttriDet}
\end{figure*}

\subsection{AttriDet: WBC detection with attributes}

Our objective is to augment existing object detectors with morphological attribute prediction modules. 
Let $\mathcal{B}$ be the feature extractor part of the object detector $\mathcal{F}$ shown in Figure \ref{fig:AttriDet}.
To capture the structure and low-level features that are necessary for morphological property recognition, we extract $N$ feature maps $\{S_j\}_{j=1}^{N}$ from the initial layers. 
For  $I_i\in \mathbb{R}^{H \times W}$ as an input image, assume  $j^{th}$ feature map as $S_j \in \mathbb{R}^{H_j \times W_j \times C_j}$.
To extract features of $k^{th}$ object using the ground-truth bounding-box $\mathbf{b}_{i,k} = \{x_k, y_k, w_k, h_k\}$ during training and predicted bounding box $\hat{\mathbf{b}}_{i,k}$ during the testing, bounding box is normalized per the size of feature map $S_j$  and image size. The normalized bounding box for $\mathbf{b}_{i,k}$ is given by:
    \begin{equation}
    \label{eq1}
    \mathbf{b}^{S_j,k}_{\text{norm}} = \left( \frac{x_k}{W_i} W_j, \frac{y_k}{H_i} H_j, \frac{w_k}{W_i} W_j, \frac{h_k}{H_i} H_j \right).
    \end{equation}
    
The normalized bounding box \( \mathbf{b}_{\text{norm}} \) is used to extract features from $S_j$ followed by ROIAlign \cite{he2017mask} to equalize the dimensions and properly align the extracted feature. 
Let \( \mathcal{R}\mathcal{O}\mathcal{I} \) denote the ROIAlign operation.
\begin{equation}
\mathcal{F}_{S_j,k} = \mathcal{R}\mathcal{O}\mathcal{I}(S_j, \mathbf{b}_{\text{norm}}^{S_j}) \quad \text{for } j \in \{1, 2\}.
\end{equation}
 
In our case, dimensions of $\mathcal{F}_{S_1,k}$ and $\mathcal{F}_{S_2,k}$ are 24$\times$30$\times$128 and 24$\times$30$\times$256, respectively. Finally, multi-scale features are fused {\(\mathcal{C_F} \)=\(\mathcal{F}_{S_{1,k}}\)\(\oplus \)  \(  \mathcal{F}_{S_{2,k}} \)}  and then fed into the AttriHead to effectively learn the relevant WBC attributes. 
In AttriHead, three convolutional layers are followed by max pooling layers, shown in Figure \ref{fig:AttriDet} (a). The output is then flattened into a d-dimensional vector and passed through a fully connected (FC) network with three layers configured as (1024, 256, 6) units. The training of AttriHead is done using multilabel binary class loss via asymmetric loss \cite{ridnik2021asymmetric}. Once trained, the AttriHead predicts the six-dimensional vectors   $\hat{\mathcal{M}}^{1 \times 6}$, where each dimension represents the state of the attribute. The details of the morphological attributes are provided in Section \ref{sub:morphology_detail}. For the MTL,  in addition to the AttriHead, we also train the YOLOv5 for WBC detection, utilizing the following loss functions with the standard settings of the detector:
\begin{equation}
\mathcal{L}_{total} = \sum_{l=1}^{N} \left( \mathcal{L}_{obj}^l + \mathcal{L}_{cls}^l + \mathcal{L}_{bbox}^l\right) + \mathcal{L}_{mor},
\end{equation}
where $l$ denotes the output prediction layers,  \(\mathcal{L}_{total}\) represents the final loss of the object detector, which is composed of several components: \(\mathcal{L}_{\text{cls}}\) represents the object classification loss, \(\mathcal{L}_{\text{bbox}}\) denotes the bounding box regression loss,  \(\mathcal{L}_{\text{obj}}\) is the loss associated with objectness and $\mathcal{L}_{mor}$ for the attribute classification.  Overall, the proposed AttriDet not only detects WBCs but also provides detailed morphology in terms of attributes.

 Attridet collectively detects cells and predicts their morphology, providing diagnostic reasoning. For each PBF image, these predictions are recorded in a morphology bank, and a blood film-level description is updated in the `Masked Text' based on the most frequently occurring WBC type and morphology, as recommended by hematologists. An example of an updated text-based description is shown in Fig. \ref{fig:AttriDet} (b). We presented the text generated from AttriDet’s predictions to hematologists, who recognized its potential as a valuable second-opinion tool for leukemia prognosis. Furthermore, AttriDet not only predicts cell-associated attributes but also improves the detector's performance, raising the mAP from 26.3 to 28.2, as shown in Table \ref{tab:Attridet_results}.

\begin{figure*}[!h]
 \centering
    {\includegraphics[width=1\linewidth]{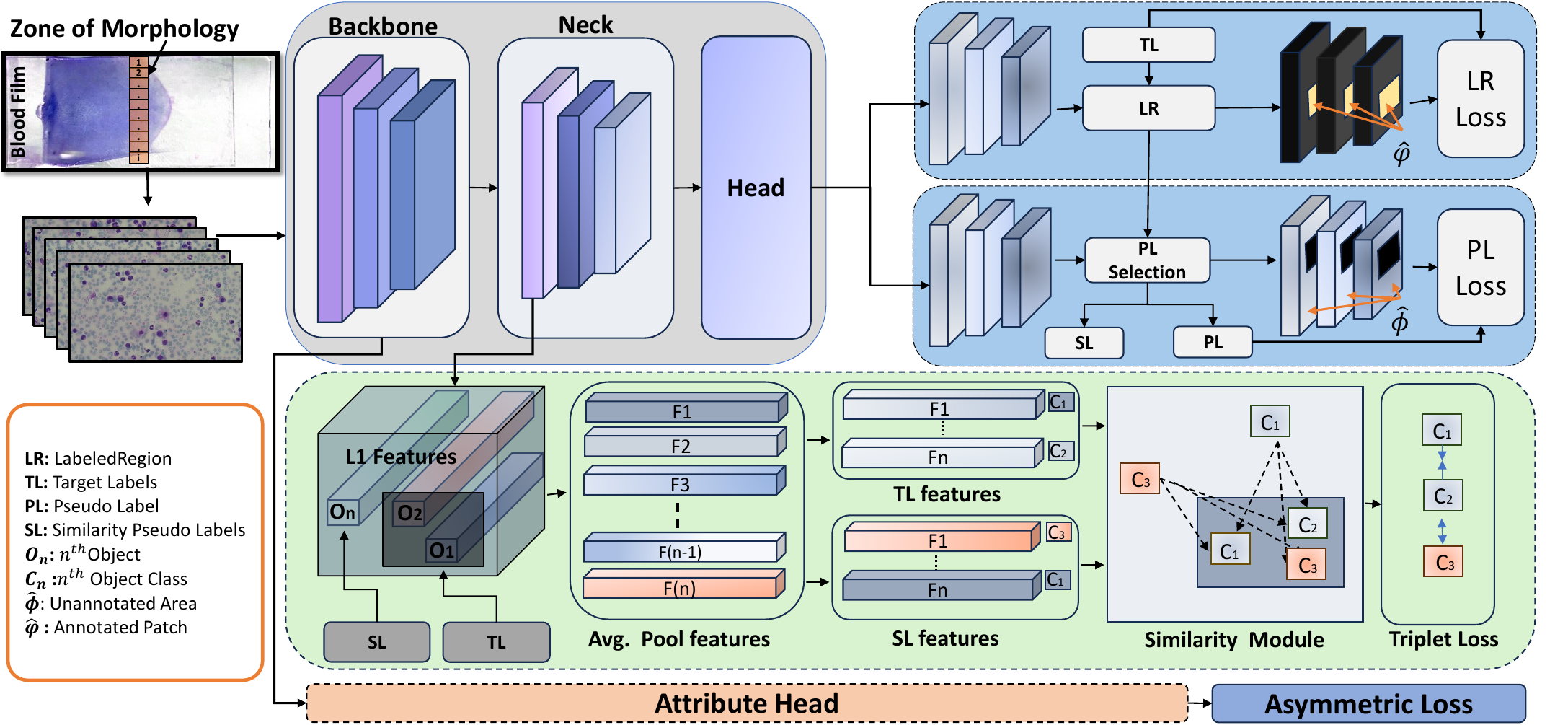}}
    \caption{
 {Architecture of SLA-Det: The proposed framework detects sparsely annotated white blood cells (WBCs) and morphological attributes in 40× images, using a backbone for feature extraction and an AttriHead module. Labeled regions (LR) optimize predictions via labeled region loss against ground-truth labels (TL), while unlabeled regions employ pseudo-labels (PL, high confidence) and similarity pseudo-labels (SL) with triplet loss to align class-based features. Training combines labeled regions, pseudo-labels, morphology-attributes, and triplet losses, enabling robust learning for accurate detection in sparse annotation scenarios.}}
    \label{fig:SLA_Det}
\end{figure*}
\subsection {SLA-Det: WBC detection with attributes using sparse annotation}
Due to annotation cost, images in the training set of sparse-leukemiaAttri are sparsely annotated. Since many object detectors (e.g.  Yolov5) use the IoU of the predicted bounding box against ground truth to identify the true positives and calculate the object loss, training with sparse annotations negatively impacts performance because the ground truth data restricts the model's ability to learn effective features, causing it to struggle with differentiating between background and objects of interest. This can increase false negatives, as the model may overlook unannotated objects.
To overcome these limitations, sparse-annotation-based object detectors,  such as SparseDet \cite{suri2023sparsedet} and Co-mining \cite{wang2021co},  have been proposed. 
Usually, these methods utilize a few bounding-box annotations of the objects (with many objects present but not annotated), without much information about the background. 
However, the scenario at hand is slightly more informative and requires redesigning the solution.
In our case, we have annotated regions, providing us with knowledge of where objects are present and where they are absent within those areas. Outside that region, we have no information about objects and backgrounds. 
For this, we propose a new sparse annotation-based object detector of leukemia, called sparse leukemia attribute detector (SLA-Det) is proposed. \\
Let $\mathcal{D_S} = \{(I_{i},  \{\mathbf{z}_{i,k}\}_{k=1}^{n_i},P_i)\}_{i=1 }^{N_D}$
be that sparsely annotated dataset.
Similar to $\mathcal{D_A}$, each annotation $z_i^k=(\mathbf{b}_{i,k}, \mathbf{c}_{i,k}, \mathbf{m}_{i,k})$ consists of bounding box $\mathbf{b}_{i,k}$, corresponding class-label $\mathbf{c}_{i,k} \in \{1, \dots, C\}$ and $\mathbf{m}_{i,k}$ is the set of morphological attributes associated with the $k^{th}$ object in the $i^{th}$ image. 
With each image $I_i$, we have associated a patch ${P}_i$, indicating a fully annotated region. Note that this region is about 20\% of the image  (as shown in Figure~\ref{fig:Sparse-LeukemiaAttri}). 
 Let $\mathcal{F}(I_i)=\{\hat{\mathbf{z}}_{i,k}\}_{k=1}^{n_i}$ denote the object detector that generates $n_i$ predictions for the image $I_i$. 
 The predicted output could be separated into non-intersecting sets $\hat{\phi}_i$ and $\hat{\varphi}_i$.
 Where $\hat{\phi}_i = \{\text{set of all~}  \hat{\mathbf{z}}_{i,k}~\text{that do not lie inside $P_i$} \}$ and 
 $\hat{\varphi}_i = \{\text{set of all~}  \hat{\mathbf{z}}_{i,k}~\text{that lie inside $P_i$} \}$.
\subsubsection{Training on Annotated Regions }
For the $\hat{\varphi}_i$ that lies inside the annotated patch $P_i$,  we can leverage the ground truth to identify correct detections and penalize any detections that occur over the background.
Let $M_i$ be the mask generated using $P_i$, such that it's zero outside the $P_i$ and 1 inside.\\ 
\noindent\textbf{Handling multiple prediction layers:}
 Several object detectors ( \cite{qiao2021detectors,lin2017feature,tan2020efficientdet}) generate predictions at multiple levels / layers $\{l^v\}_{v=1}^{N}$, to detect objects at multiple scales. 
 In order to penalize the objectness score assigned to the background region, the location of the annotated patch has to be adjusted to these layers. 
 We  scale ${P}_i$ for $l^v$ by applying normalization (Equation \ref{eq1}) creating  ${P}^{v}_{i,\text{norm}}$.
For each level, we separate the predictions, such that $\forall_v \hat{\varphi}_i^v = \{\text{set of all~}  \hat{\mathbf{z}}_{i,k} \text{that lie inside $P_i^v$} \}$.
\noindent\textbf{Labeled region loss: }
To fully utilize the information in the annotated patch, the objectness loss has to be backpropagated. 
Since objectness is different at different layers (as each layer predicts objects of different sizes), we compute the masked objectness loss for each layer using the prediction of $\hat{\varphi}_i^v$ that lies within the annotated patch. This ensures that the loss calculation focuses only on relevant regions, allowing the model to better learn from the annotated objects while ignoring irrelevant background information. Final loss for annotated region is Eq.~(\ref{eq:LossSparseAnn}):
\begin{equation} 
\label{eq:LossSparseAnn}
\mathcal{L}_{LR} = \sum_{v=1}^{N} \left( \mathcal{L}_{obj}(\hat{\varphi}_i^v) + \mathcal{L}_{cls}(\hat{\varphi}_i^v) + \mathcal{L}_{bbox}(\hat{\varphi}_i^v) \right), 
\end{equation}
\subsubsection{Training on non-Annotated Regions}
To address the lack of annotations in unlabeled regions, we rely on a self-training approach by selecting pseudo-labels from the predictions in the non-annotated region $\hat{\phi}_i$.

\noindent\textbf{Pseudo-Label filtering:} Since predictions from the not-properly trained model are noisy and might be incorrect, we design filtering criteria to select the pseudo-labels from the predictions.
First, we filter out predictions, $\mathbf{z}^*_{i,k}$,  that have predicted confidence score,  \( {o} ^*_{i,k} \) below $\mathbf{t}_{0}$. 
 This effectively means that only predictions from unlabeled regions with a high enough confidence, considering both the class probability and object score, are retained for further processing. 
Secondly, we remove all object predictions with an area smaller than the smallest object (\(\text{Area}(\hat{b}_{i,k}) > \text{Area}_{\text{min}}\)) in the ground-truth in the training set. Thirdly, in the remaining set, for any prediction with confidence score \(\hat{{o}}^*_{i,k}\) above \(\mathbf{t}_{1}\), we compute entropy, allowing us to focus on the most distinguished class probabilities \cite{grandvalet2004semi}:

\begin{equation}
\mathcal{E}_{i,k} = - \sum_{c' \in \mathcal{C}}  \hat{\mathbf{p}}_{c'}  \log_2 \hat{\mathbf{p}}_{c'},
\end{equation}  
where \( \mathcal{C} \) is the set of all class probabilities of \( {c}^{*}_{i,k} \), and \( \hat{\mathbf{p}}_{c'} \) is the probability of the outcome \( c' \). In summary,

\begin{equation}
\begin{split}
\label{eq12}
D_{i}^{pseudo} = \{ \hat{\mathbf{z}}_{i,k} \in \hat{\phi}_i \mid \text{Area}(\hat{b}_{i,k}) > \text {Area}_{\text{min}}) ~\&\\ \hat{c}_{i,k}>\mathbf{t}_{1}~\& ~\mathcal{E}_{i,k}<\mathbf{t}_{2} \},
\end{split}
\end{equation}
where $\mathbf{t}_{2}$ denotes the entropy threshold.
The above-selected pseudo-labels could not be used to reject the background as they only tell where the object is, but not where an object could not be. Therefore, we need to mask out the object loss for the regions outside the pseudo-label. 
For $\hat{\mathbf{z}}_{i,k} \in D_{i}^{pseudo} $,  we define the pseudo-label 
${\mathbf{z}}_i^{*k} =( \hat{{c}}_{i,k}, \hat{\mathbf{b}}_{i,k}, M_{i,k})$, where $M_{i,k}$ is the mask of image size with only object location set to 1, rest is zero. 
Let $\phi_i^{*} = \{ \mathbf{z}_i^{*k}| \hat{\mathbf{z}}_{i,k} \in D_{i}^{pseudo}$ \}. For each level, the elements in the set are normalized per the size of the level and result in sets   $\phi_i^{*v}$.
\begin{equation}
\label{eq10}
\mathcal{L}_{PL} = \sum_{v=1}^{N} \left( \mathcal{L}_{\text{obj}}(\hat{\phi}_i^v) + \mathcal{L}_{\text{cls}}(\hat{\phi}_i^v) + \mathcal{L}_{\text{bbox}}(\hat{\phi}_i^v) \right),
\end{equation}
where the $\mathcal{L}_{PL} $ represent the loss from the unlabeled region.

\noindent\textbf{Similarity based training:} 
We hypothesize that if the classes of two objects (in our case, the object class from the labeled region versus the unlabeled region) are the same, then their features should be very similar. Conversely, if the classes are not the same, their features should not be similar. Therefore, the minimum similarity between objects of the same class must be greater than the maximum similarity between objects of different classes.}
Based on this hypothesis,  we design a novel strategy to use  $\hat{\mathbf{z}}_{i,k} \in \hat{\phi}_i$ which have confidence score greater than $\mathbf{t}_{0}$ indicating they might not be background but less than $\mathbf{t}_{1}$ (indicating we are not sure of the class label).
Let $\hat{\mathbf{z}}^{s}_{i,k}$ be such predictions. 
For $\hat{\mathbf{z}}^{s}_{i,k}$, we extract the features,$\hat{f}_{i,k}$, from the initial layers of neck. 
Similarly, for each $j^{th}$ ground-truth bounding box in annotated patch ($P_i$), we extract  ${f}_{i,j}$.
To enhance adaptability, the triplet difference is determined by evaluating the feature maps of each filtered label against all ground truth feature maps within the batch:
\begin{equation}
\label{eq13}
\begin{split}
\text{Min\_Sim}(\hat{f}_{i,k}) = \min_{p \in {f}_{i}} \left( \frac{p \cdot \hat{f}_{i,k}}{\|p\| \|\hat{f}_{i,k}\|} \right),
\text{for all } \hat{c}^{s}_{i,k} = \hat{c}_{i,j}
\end{split}
\end{equation}
\begin{equation}
\label{eq13}
\begin{split}
\text{Max\_Sim}(\hat{f}_{i,k}) = \max_{p \in {f}_{i}} \left( \frac{p \cdot \hat{f}_{i,k}}{\|p\| \|\hat{f}_{i,k}\|} \right),
\text{for all } \hat{c}^{s}_{i,k} \neq \hat{c}_{i,j}
\end{split}
\end{equation}
\begin{equation}
{Tri}_{i,k} = \max(0, \text{Min\_Sim}(\hat{f}_{i,k}) - (\text{Max\_Sim}(\hat{f}_{i,k}) + 0.05)).
\end{equation}
The total triplet  loss is computed as: 
\begin{equation}
\label{eq15}
\mathcal{L}_{Tri} = \frac{1}{m}\sum_{k=1}^{m} {Tri}({ \hat{f}}_{i,k}, {{f}}_{i,j}),
\end{equation}
where \(m\) represent the numbers of object features in ${ \hat{f}}_{i,k}$. The final loss function for learning to detect leukemia WBCs in sparsely annotated settings is as follows:
\begin{equation}
\label{eq:final}
\mathcal{L}_{SL} = \mathcal{L}_{{LR}} + \mathcal{L}_{{PL}} + \mathcal{L}_{Tri}.
\end{equation}
Similar to AttriDet, SLA-Det is built upon the YOLOv5 object detector (Fig. \ref{fig:SLA_Det}); however, the presented methodology for the AttriHead is adopted for the attribute prediction with the same setting and only for the region where the ground truth is available. Finally, to learn to predict attributes (using AttriHead) along with WBC detections, we employ the following loss.
\begin{equation}
\label{eq:final}
\mathcal{L}_{SLA} = \mathcal{L}_{{LR}} + \mathcal{L}_{{PL}} + \mathcal{L}_{Tri} + \mathcal{L}_{mor} 
\end{equation}
 
\section{Experiments and Results}

\subsection{Experiments  on \datasetA}

   \noindent\textbf{Implementation details:}
 For AttriHead, dropout regularization (20\%) is applied between FC layers, with Leaky ReLU activation used for both convolutional and FC layers. AttriDet is trained on the LeukemiaAttri dataset with specified augmentations and loss functions. We set the IoU threshold at 0.20 and a mosaic augmentation probability of 1.0. The optimizer settings include a weight decay of 0.0005, momentum of 0.937, and warmup biases of 0.1. Training is conducted for 100 epochs with a batch size of 4 on an NVIDIA GTX 1080 GPU, using a learning rate: of (0.01 to 0.001) and a 3-epoch warmup.

\noindent\textbf{WBC detection: }To evaluate WBC detection performance on \datasetA, we perform experiments with several well-known object detectors including 
Sparse R-CNN \cite{sun2021sparse}, FCOS \cite{tian2019fcos}, DINO\cite{zhang2022dino}, and YOLOv5x \cite{ultralytics2021yolov5}.
 We have performed detailed experiments on the dataset collected using a low-cost microscope (LCM) and a high-cost microscope (HCM), employing mobile cameras (C1) and special-purpose microscopic cameras (HD1500T for HCM and MZZCAT 5MP for LCM - C2) on 40x and 100x magnifications.
 The experimental results are shown in  Table \ref{tab:Baseline_results}. The results indicate that the best WBC detection accuracy ({mAP@$_{50-95}$} of 26.3 and {mAP@$_{50}$} of 44.2) is achieved on the data collected using HCM  at 100x employing C2 camera using YOLOv5x. Similarly, when comparing on 40x, YOLOv5x outperforms ($\text{mAP}_{50\text{-}95}$ of 20.1 and $\text{mAP}_{50}$ of 37.3) other objection detection methods. A similar pattern can be observed for other subsets as well. We believe this is due to the YOLOV5 robust feature learning for different sizes of cells in the LeukemiaAttri dataset. Given that YOLOv5x demonstrated superior performance across both the HCM and LCM subsets of microscope and mobile camera data, we extend this for attribute prediction.
\begin{table}[h]
\caption{Object Detection baselines results (\(mAP_{@50}\)) on multipal subset of \datasetA~dataset. The C2 camera represents the HD1500T camera for HCM, while the C2* camera corresponds to the ZZCAT 5MP camera for LCM. The C1 mobile camera, used for both, is the Honor 9x Lite.} 
\label{tab:Baseline_results}
\centering
\resizebox{\columnwidth}{!}{%
\begin{tabular}{|c|c|c|c|c|c|c|c|c|c|}
\hline
 &\multicolumn{4}{|c|}{High cost Microscope}&\multicolumn{4}{|c|}{Low cost Microscope} \\ \cline{2-9}
 Methods & \multicolumn{2}{c|}{100x} &\multicolumn{2}{c|}{40x}& \multicolumn{2}{c|}{100x} &\multicolumn{2}{c|}{40x}\\
\cline{2-9}
 &C1 & C2 & C1 & C2&C1 & C2* & C1 & C2*  \\  
\hline
{Sparse R-CNN}&29.6&36.7&27.0&32.7&32.6&25.9&26.4&33.9\\
{FCOS}&31.8&40.6&24.8&32.7&33.9&34.3&31.2&28.5\\
{DINO}&33.8&43.7&\textbf{36.1}&36.9&34.3&\textbf{38.2}&\textbf{31.4}&\textbf{36.6}\\
{YOLOv5x}&\textbf{38.8}&\textbf{44.2}&\textbf{36.1}&\textbf{37.3}&\textbf{39.5}&38.1&29.7&34.9\\


\hline
\end{tabular}
}
\end{table}

\noindent\textbf{WBC detection with attribute prediction:} Table. \ref{tab:Attridet_results} demonstrate results  {of} our proposed AttriDet for attributes prediction. In addition, we demonstrate the improved WBC detection of AttriDet (last column) as compared to standard YOLOv5. We credit the enhanced WBC results to robust feature learning using the attributes head. Findings from CBM \cite{keser2023interpretable}  and AttriDet demonstrate superior performance in predicting WBC types and their attributes. However, detecting nucleus and cytoplasmic vacuoles remains a significant challenge. 
\begin{table}[!htbp]
\centering
\caption{Results on HCM\_100x\_C2 test set.WBC (mAP@$_{50\text{-}95}$),  Attributes (F1): NC (Nuclear Chromatin), NS (Nuclear Shape), N (Nucleus), C (Cytoplasm), CB (Cytoplasmic Basophilia), CV (Cytoplasmic Vacuoles).}
\label{tab:Attridet_results}
\resizebox{\columnwidth}{!}{%
\begin{tabular}{|l|cccccc|c|}
\hline
Method & NC & NS & N & C & CB & CV & WBC  \\
\hline
CBM & 21.9 & \textbf{96.2} & 41.8 & 77.2 & 70.2 & 3.3 & 27.6 \\
AttriDet  & \textbf{73.9} & 95.9 & \textbf{54.3} & \textbf{89.7} & \textbf{83.6} & \textbf{29.1} & \textbf{28.2} \\
\hline
\end{tabular}
}
\end{table}
\\

{\noindent\textbf{Ablation of attrihead}. The ablation of the AttriHead architecture could offer useful insights. Our focus was on developing a lightweight model tailored to real-world clinical constraints and limited computational resources. The chosen design, guided by empirical observations, balances performance and efficiency, as demonstrated by the strong results reported in Table \ref{tab:attrihead_ablation} under low resource settings.}
\begin{table}[!htbp]
\centering
\caption{{Ablation study of activation functions and architectural configurations of Attrihead on attribute-level F1 scores.}}
\resizebox{\columnwidth}{!}{%
\begin{tabular}{|c|c|c|cccccc|}
\hline
{{Activation}} & {{FC}} & {{Conv}} & {{NC}} & {{NS}} & {{N}} & {{C}} & {{CB}} & {{CV}} \\
& {{Layers}} & {{Layers}} &  &  &  &  &  & \\
\hline
{LeakyReLU} & {3} & {3} & {\textbf{73.9}} & {95.9} & {\textbf{54.3}} & {\textbf{89.7}} & {83.6} & {\textbf{29.1}} \\
{ReLU}      & {3} & {3} & {64.1}   & {96.2}   & {47.8}   & {88.1}   & {83.8}   & {11.2} \\
{LeakyReLU} & {3} & {2} & {65.8} & {\textbf{96.3}} & {42.8} & {87.7} & {84.2} & {04.7} \\
{LeakyReLU} & {2} & {3} & {63.8}   & {96.0}   & {50.2}   & {\textbf{89.7}}   & {83.5}   & {2.3} \\
{LeakyReLU} & {2} & {2} & {65.0}   & {\textbf{96.3}}   & {45.6}   & {88.1}   & {\textbf{84.3}}   & {10.8} \\
\hline

\end{tabular}
}
\label{tab:attrihead_ablation}
\end{table}\\
\noindent\textbf{Unsupervised domain adaptation based WBC detection:} \\
The LeukemiaAttri dataset includes 12 subsets originating from distinct domains, captured via HCM and LCM. These subsets exhibit significant domain shifts, as illustrated in Fig. \ref{fig:Leukemiaattri_data} (a), positioning our dataset as a valuable benchmark for domain adaptation studies. It is worth noting that LCM images often suffer from poor quality, discouraging hematologists from providing annotations due to the labor-intensive and error-prone nature of the process. As a solution, UDA approaches enable the training of object detectors on high-quality, well-annotated HCM images and facilitate their application to LCM images. 
\begin{table}[!htbp]
\caption{Domain adaptation results for object detection on the \datasetA~dataset}
\label{tab:UDA_results}
\centering
\begin{tabular}{|c|c|c|}
\hline
{Method} & {mAP@${_{50-95}}$}& {mAP@${_{50}}$} \\
\hline
YOLOv5&11.0&25.5\\ 
DACA  &12.6 &30.2\\ 
ConfMix  &12.6&33.5\\ \hline
\end{tabular}
\end{table}

To establish the UDA baselines, we evaluated UDA baselines using ConfMix \cite{sun2021sparse} and DACA \cite{mekhalfi2023detect} on high-resolution subsets (HCM\_100x\_C2 as source and LCM\_100x\_C2 as target).  Table \ref{tab:UDA_results} demonstrates that YOLOv5x (source only) achieved 25.5 mAP@${_{50}}$, while UDA methods improved performance to 33.5 (ConfMix) and 30.2 (DACA), highlighting the significant domain shift and complexity of our dataset.
\setlength{\tabcolsep}{19.5pt}
\begin{table}[h]
\caption{Comparative results on \datasetB}
\label{tab:SLA_Detetion_results}
\centering
\resizebox{\columnwidth}{!}{%
\begin{tabular}{|c|c|c|}
\hline
{Method} & {mAP@$_{50-95}$}& {mAP@$_{50}$} \\
\hline
  SparseDet&6.1 &10.4  \\
  Co-mining &8.6&15.3 \\
  SLA-Det (Ours)&\textbf{27.2}&\textbf{45.1}\\
 \hline
\end{tabular}
}
\end{table}

\subsection{Experiments  on \datasetB}
\noindent\textbf{Implementation details:}
For the SLA-Det implementation, we set the values as follows: $\mathbf{t}_{0}: 0.70$, $\mathbf{t}_{1}: 0.95$, $\mathbf{t}_{2}$: 2.6. Additionally, the batch size is set to 4, and both pseudo and triplet losses are used with a weight of 0.1. During training for WBC detection, the triplet loss is utilized for 40 epochs, while the other loss functions are applied for the entire 50 epochs. In the case of detection with attribute prediction, the triplet loss is activated for 10 epochs, while the others are for 30. {In all of the experiments, we used the final model weights (i.e., the weights from the last training epoch) for evaluation. The same dataset split and annotations are used for our approach and the baseline methods.

In our setting, approximately 20\%  regions per image are annotated, and the rest remain unannotated. For the SOTA methods, we strictly followed their official implementation and default hyperparameters as provided in their publicly available repositories.}\\

\noindent\textbf{WBC detection: } We compare the proposed SLA-Det with recent state-of-the-art sparse learning methods \cite{suri2023sparsedet,wang2021co}. {The results in Table \ref{tab:SLA_Detetion_results} demonstrate that our approach outperforms existing methods, achieving a notably higher mAP.} 
The superior results of our approach are due to effectively utilizing annotated information through labeled region masking, strategic use of pseudo-labels, and integrating cell similarity.    

\setlength{\tabcolsep}{7pt} 
\begin{table}[!htbp]
\caption{{Ablation results of SLA-Det on \datasetB for detection only,  we conduct multiple runs (N=3) for each ablation setting and compute the mean and standard deviation of both {mAP@$_{50-95}$} and {mAP@$_{50}$}.}
}
\label{tab:ablation_studies}
\centering

\begin{tabular}{|c|c|c|c|c|}
\hline
\cline{2-4}
{Labeled} & {Pseudo} & {Triplet} & {mAP@$_{50-95}$} & {mAP@$_{50}$} \\
{Region} & {label} &  & {(±std)} & {(±std)} \\
\hline
{\ding{52}} & {\ding{56}} & {\ding{56}} & {$24.6_{(\pm0.5)}$} & {$40.6_{(\pm0.4)}$} \\
{\ding{52}} & {\ding{52}} & {\ding{56}} & {$24.8_{(\pm0.2)}$} & {$41.3_{(\pm0.4)}$} \\
{\ding{52}} & {\ding{52}} & {\ding{52}} & {$\mathbf{26.4}_{(\pm0.4)}$} & {$\mathbf{44.6}_{(\pm0.5)}$} \\
\hline
\end{tabular}

\end{table}

\noindent\textbf{Ablation studies:}
Our method contains several components. We analyze the components of the SLA-Det approach for WBC detection to verify their effectiveness. Results in Table \ref{tab:ablation_studies} show that each component contributes to the final accuracy, with LR loss improving detection mAP by 3.1\%. PL loss achieves an additional 0.7\% improvement, and applying triplet loss results in a significant 3.6\% increase. Our results demonstrate that each component of our approach contributes effectively.

\setlength{\tabcolsep}{4pt}
\begin{table*}[!b]
\center

\caption{{Ablation study results: F1 Score of SLA-Det on \datasetB. Attributes are the same as in Table \ref{tab:Attridet_results}. Multiple runs (N=3) are conducted for each ablation setting and compute the mean and standard deviation of {mAP@$_{50-95}$}.}}
\label{tab:SLA_Ablaition}
\begin{tabular}{|c|cccccc|c|}
\hline
{Method} & {NC} & {NS} & {N} & {C} &{CB} &{CV}&{mAP@$_{50-95}$}\\
\hline
{SLA-Det*} & {$70.8 _{(\pm 0.6)}$} & {$95.5 _{(\pm 0.07)}$} & {$50.9 _{(\pm 0.07)}$} & {$88.3 _{(\pm 0.7)}$} & {$83.9 _{(\pm 0.6)}$} &  {$ 14.7_{(\pm 0.9)}$} & {$26.4 _{(\pm 0.5)}$} \\

{SLA-Det**} & {$69.1 _{(\pm 1.9)}$} & {$97.5 _{(\pm 0.2)}$} & {$50.4 _{(\pm 1.1)}$} & {$87.8 _{(\pm 0.7)}$} & {$83.6 _{(\pm 0.7)}$} &  {$ 11.9_{(\pm 3.1)}$} & {$26.9 _{(\pm 0.7)}$} \\

{SLA-Det***} & {$\textbf{71.1} _{(\pm 1.2)}$} & {$\textbf{97.5} _{(\pm 0.1)}$} & {$\textbf{51.2} _{(\pm 0.8)}$} & {$\textbf{88.0} _{(\pm 0.3)}$} & {$\textbf{84.2} _{(\pm 0.2)}$} & {$\textbf{13.2} _{(\pm 1)}$} & {$\textbf{28.2} _{(\pm 0.7)}$} \\
\hline


\end{tabular}
\end{table*}

We have also tried different straightforward methods to address the sparse dataset problem. In the first experiment, the model is trained on cropped, annotated patches from \datasetB, while during testing, each image is divided into overlapping patches. In the second experiment, the model is trained on the full image with unannotated regions masked out, and the entire image is used during testing. Both experiments yielded $mAP@{_{50}}$ scores of 37.7\% and 37.6\%, respectively, which are noticeably lower than the 45.1\% mAP@${_{50}}$ maximum achieved by our proposed approach.\\
\noindent\textbf{WBC detection with attribute prediction :} 
Table \ref{tab:SLA_Ablaition} presents the results of our SLA-Det model for various prediction of attributes. In this table, SLA-Det* denotes the AttriHead with only the LR loss. SLA-Det** includes both LR and PL losses with the AttriHead, representing the second phase of the ablation study. Finally, SLA-Det*** shows the results from the final phase of the ablation study, incorporating all losses (LR+PL+Triplet) with the AttriHead.\\
{\noindent\textbf{Ablation of thresholds :} 
The thresholds $t_{0}$, $t_{1}$, and $t_{2}$ used in the pseudo-label filtering step of SLA-Det play a critical role in influencing detection performance. These values were selected empirically based on preliminary experiments. To quantify their impact, an ablation was conducted by varying each threshold within a reasonable range. The results, presented in Table \ref{tab:threshold_ablation}, demonstrate that our method is robust to slight changes in these parameters.}
\begin{table}[!htbp]
\centering
\caption{ {Ablation of pseudo-label thresholds ($t_0$, $t_1$, $t_2$) on SLA-Det performance.}}
\resizebox{\columnwidth}{!}{%
\begin{tabular}{ccc|cccccc|cc}
\hline
{\textbf{$t_0$}} & {\textbf{$t_1$}} & {\textbf{$t_2$}} & {NC} & {NS} & {N} & {C} & {CB} & {CV} & {$mAP@_{50-95}$} & {$mAP@_{50}$} \\
\hline
{0.70} & {0.95} & {2.6} & {70.9} & {\textbf{97.7}} & {51.3} & {87.9} & {84.3} & {12.1} & {\textbf{29.0}} & {\textbf{46.4}} \\
{0.70} & {0.93} & {2.6} & {68.4} & {{97.6}} & {49.9} & {\textbf{89.5}} & {83.3} & {\textbf{13.2}} & {{25.5}} & {{41.5}} \\
{0.70} & {0.90} & {2.6} & {70.9} & {\textbf{97.7}} & {\textbf{52.5}} & {88.6} & {83.4} & {11.6} & {19.6} & {33.2} \\
{0.60} & {0.95} & {2.6} & {\textbf{71.6}} & {97.3} & {52.1} & {\textbf{89.5}} & {\textbf{84.8}} & {{13.1}} & {26.9} & {44.3} \\
{0.70} & {0.95} & {2.5} & {70.8} & {97.1} & {51.7} & {89.1} & {83.1} & {10.3} & {25.7} & {42.4} \\
{0.70} & {0.95} & {2.7} & {70.3} & {97.6} & {50.5} & {89.0} & {84.0} & {8.0}  & {26.4} & {42.9} \\
\hline

\end{tabular}
}
\label{tab:threshold_ablation}
\end{table} \\

{\noindent\textbf{Discussion:} The model's difficulty in accurately detecting the `Cytoplasmic Vacuoles' attribute is primarily due to two main factors. First, there is a significant class imbalance in the dataset (see Table \ref{tab:classes_details}). Second, cytoplasmic vacuoles are morphologically linked to cell swelling and early necrotic changes, which are often difficult to distinguish under variable staining and imaging conditions \cite{chiang2018automated}. Their nonspecific appearance across multiple pathological states and overlap with other intracellular features further adds to annotation and detection challenges.}
{In routine clinical workflows, clinicians typically perform a quick differential count in a few minutes; however, detailed morphological assessment requires expertise and is time-consuming. A model like SLA-Det, trained on a dataset with annotated morphological attributes, can assist hematologists in real time during inference by providing cell type and morphology insights as a second opinion. While only $\approx 20\%$ of each image is annotated for training to reduce annotation burden, the model is designed to operate on the entire field of view (FoV) during inference. With an average processing time of $\approx 0.29$ seconds per image ($\approx 3.45$ frames per second), the model operates at near real-time performance suitable for clinical workflows. 
}

\begin{figure*}[!t]
 \centering
\includegraphics[width=1\linewidth]{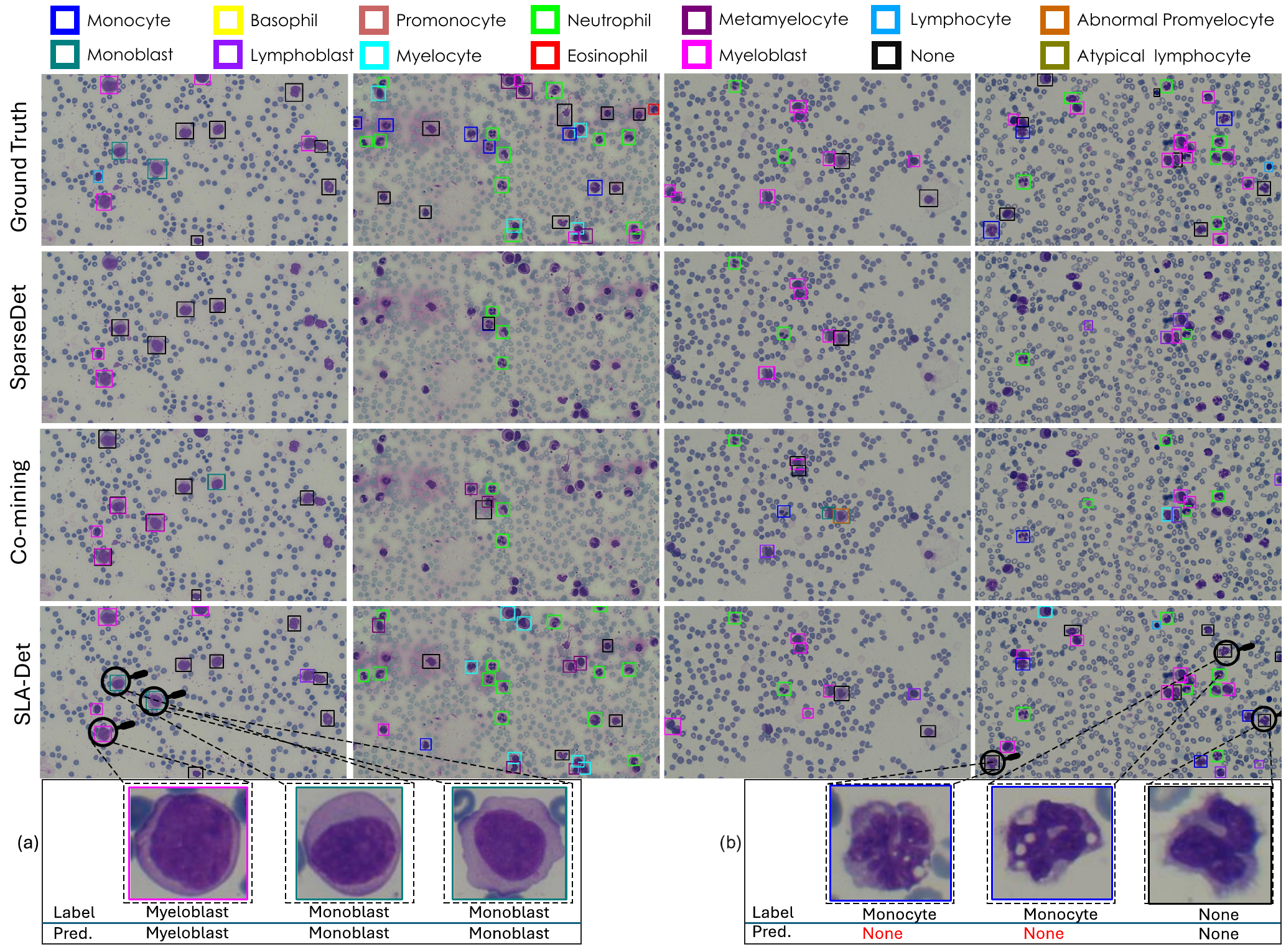}
    \caption{ {Qualitative comparison of different sparsely annotated approaches, including SparseDet \cite{suri2023sparsedet} and Co-mining \cite{wang2021co}, and SLA-Det. Previous methods struggle to accurately localize all WBCs (columns 2 and 4), whereas the proposed approach effectively detects WBCs with improved classification. A zoomed-in view highlights (a) successful detection cases and (b) misclassified cases labeled as None (details are given in the text). }}
    \label{fig:qualatative}
\end{figure*}
\noindent\textbf{Qualitative results: }
The qualitative comparison of the proposed approach and existing sparse object detectors \cite{wang2021co,suri2023sparsedet} is shown in Figure \ref{fig:qualatative}. {The results suggest that the proposed method offers improved localization and classification performance compared to competitive approaches.} Figure \ref{fig:qualatative} (a) and Figure \ref{fig:qualatative} (b) show the zoomed-in version of some of the cells.  In Figure \ref{fig:qualatative} (a), all three cells belong to the same myeloid lineage and exhibit nearly identical morphological features. The primary difference between myeloblasts and monoblasts is the nucleus's position within the cytoplasm: monoblasts have a nucleus surrounded by a more abundant and vacuolated cytoplasm, while myeloblasts have a more distinct nuclear-cytoplasmic boundary. Despite these minor differences, the detector effectively identifies both cell types. Figure \ref{fig:qualatative} (b) demonstrates a failure case of our approach, where the shapes of the cell nucleus and cytoplasm are highly irregular. The first two cells are labeled as `Monocyte' and the third cell is labeled as `None'. 
We believe that the minor staining effects might be the reason for confusing the detector, potentially resulting in incorrect classifications. 
\section{Conclusion}
We present a Large Leukemia Dataset (LLD) consisting of datasets named \datasetA~ and \datasetB.  
The \datasetA~ contains 12 subsets collected  by using two different quality microscopes with multiple cameras at
different resolutions (10x, 40x, and 100x). 
There are 2.4k annotated images per resolution (in both HCM and LCM), with 10.3K cells annotated for 14 WBC types and 7 morphological properties.
Based on morphological information, we have provided an AttriDet method to detect the WBC type with its morphological attributes. The \datasetB~is introduced to tackle the real-world challenges of sparse annotation in microscopy images. To establish the baseline solutions, we propose SLA-Det, which effectively utilizes labeled and unlabeled region information. We believe that the presented dataset and methods will advance research in explainable, robust, and generalizable leukemia detection. {Our propose
 setup ensures a clean and consistent sparse annotation structure. Real-world clinical annotations are often irregular and unstructured, with cells marked at arbitrary locations. The current SLA-Det model does not support unstructured annotation patterns. Addressing this limitation will be a focus of future work.}

\section*{\ackname} 
We extend our sincere gratitude to Dr. Asma Saadia from Central Park Medical College, Lahore, and Dr. Ghulam Rasul from Ittefaq Hospital, Lahore, for their insightful discussions. Additionally, we express our appreciation to the dataset collection team, particularly Aurang Zaib, Nimra Dilawar, Sumayya Inayat, and Ehtasham Ul Haq, for their dedicated efforts. Furthermore, we acknowledge Google for their partial funding support for this project.






\bibliography{main.bib}
\bibliographystyle{elsarticle-num-names}
\end{document}